\pdfoutput=1

\documentclass[11pt]{article}

\usepackage[final]{acl}

\usepackage{times}
\usepackage{latexsym}

\usepackage[T1]{fontenc}

\usepackage[utf8]{inputenc}

\usepackage{microtype}

\usepackage{inconsolata}

\usepackage{graphicx}

\usepackage{booktabs}
\usepackage{tabularx}
\usepackage{lipsum}

\title{Can Prompts Rewind Time for LLMs?\\Evaluating the Effectiveness of Prompted Knowledge Cutoffs}

\author{
  Xin Gao\textsuperscript{1}\thanks{\ Equal contribution.} \quad
  Ruiyi Zhang\textsuperscript{1}\footnotemark[1] \quad
  Daniel Du\textsuperscript{1} \quad
  Saurabh Mahindre\textsuperscript{2} \quad \\ \bf
  Sai Ashish Somayajula\textsuperscript{1}\footnotemark[2] \quad
  Pengtao Xie\textsuperscript{1}\thanks{\ Corresponding authors.} \\
  \\\
  \textsuperscript{1}UC San Diego
  \textsuperscript{2}SUNY Buffalo \\
  \texttt{\{xig022, ruz048, ssomayaj, p1xie\}@ucsd.edu}
}

\begin{document}
\maketitle
\begin{abstract}
Large Language Models (LLMs) are widely used for temporal prediction, but their reliance on pretraining data raises contamination concerns, as accurate predictions on pre-cutoff test data may reflect memorization rather than reasoning, leading to an overestimation of their generalization capability. With the recent emergence of prompting-based unlearning techniques, a natural question arises: Can LLMs be prompted to simulate an earlier knowledge cutoff? In this work, we investigate the capability of prompting to simulate earlier knowledge cutoff in LLMs. We construct three evaluation datasets to assess the extent to which LLMs can forget (1) direct factual knowledge, (2) semantic shifts, and (3) causally related knowledge. Results demonstrate that while prompt-based simulated knowledge cutoffs show effectiveness when directly queried with the information after that date, they struggle to induce forgetting when the forgotten content is not directly asked but causally related to the query. These findings highlight the need for more rigorous evaluation settings when applying LLMs for temporal prediction tasks. 
The full dataset and evaluation code are available at \url{https://github.com/gxx27/time_unlearn}.
\end{abstract}

\section{Introduction}

Large Language Models (LLMs) have shown strong capabilities in knowledge extraction and information processing, leading to their adoption in temporal prediction tasks such as stock forecasting and event prediction~\citep{wang2024from,yu-etal-2023-harnessing}. However, evaluating their performance on these tasks is challenging, as LLMs are pretrained on large-scale web corpora and may have seen information from the test data~\citep{dong-etal-2024-generalization}. Take the stock price prediction task as an example: typically, we train a machine learning model, such as a Random Forest (RF)~\citep{breiman2001random}, from scratch using stock prices of a company from 1960 to 2010, and evaluate its prediction performance on data from 2010 to 2015. The resulting test performance is generally reliable~\citep{10.1093/rfs/hhaa009}. However, suppose we adopt the same experimental setup but replace the RF with an LLM predictor. In that case, the test performance is no longer trustworthy, as the LLM may have already encountered the 2010–2015 stock data during pretraining. This can lead to overestimated performance and poor generalization on prediction tasks occurring after the model's actual knowledge cutoff~\citep{roberts2024to}.

\begin{figure}
    \centering
    \includegraphics[width=\linewidth]{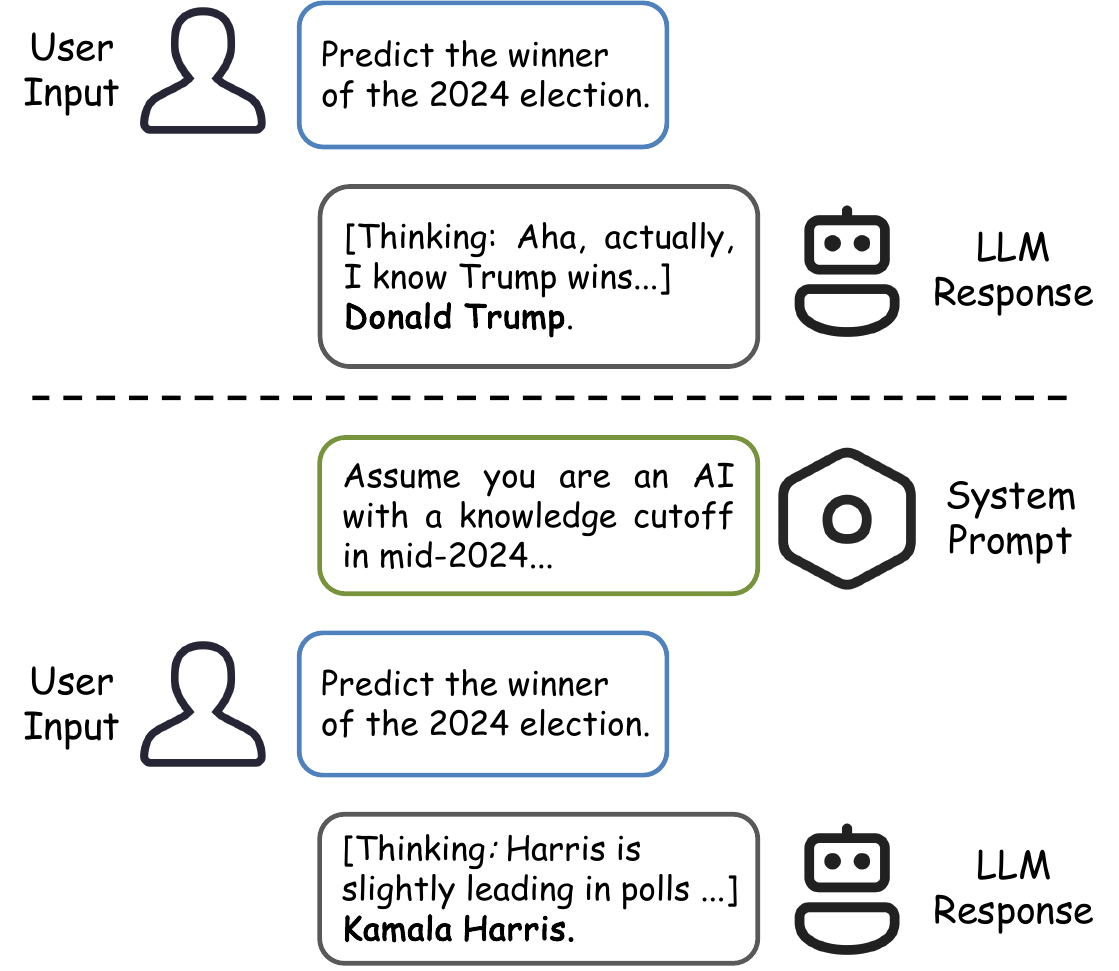}
    \caption{\textbf{Top:} The LLM answers the user’s question using memorized knowledge. \textbf{Bottom:} The LLM does not use memorized knowledge to respond, given the prompted knowledge cutoff.}
    \label{fig:motivation}
\end{figure}

\begin{figure*}
    \centering
    \includegraphics[width=\linewidth]{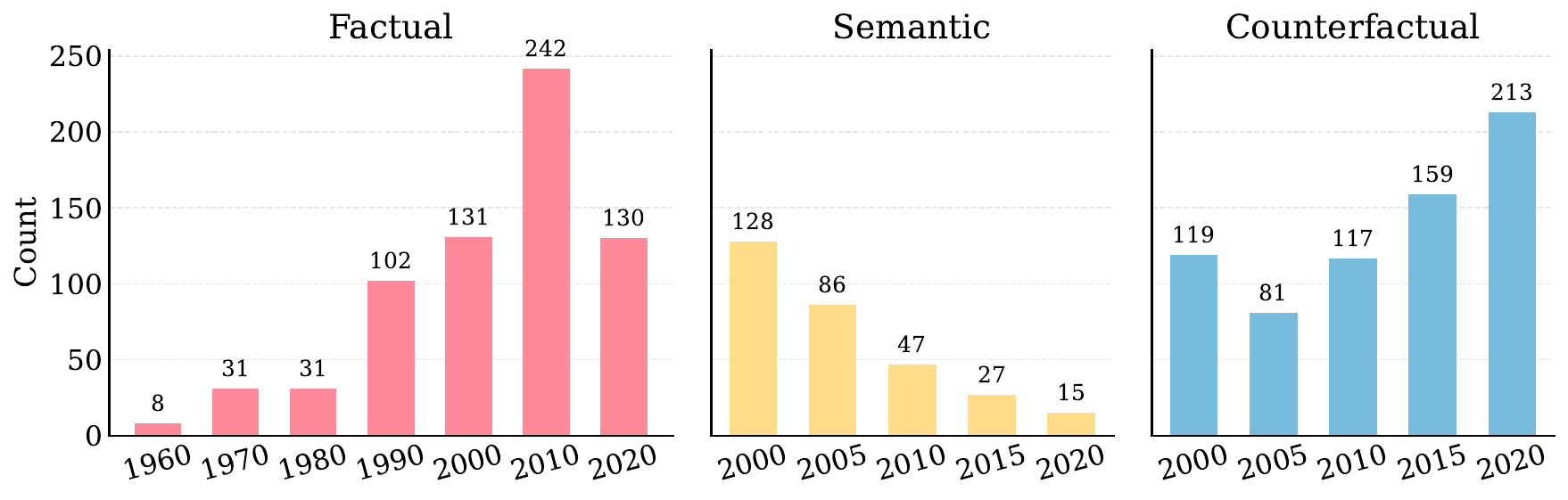}
    \caption{Distribution of data instances by year across the \emph{Factual}, \emph{Semantic}, and \emph{Counterfactual} subsets.}
    \label{fig:data-year}
\end{figure*}

Recent work on in-context unlearning has explored how LLMs can be guided to forget specific data instances or concepts through prompting alone~\citep{Pawelczyk2024incontext}. Motivated by this, we ask: \emph{Can prompting be used to adjust an LLM's knowledge cutoff, inducing it to unlearn all information beyond the cutoff date?} If so, this approach could mitigate the data contamination issue discussed earlier and enable more trustworthy evaluation, as intuitively illustrated in two examples in Figure~\ref{fig:motivation}.

To investigate this question, we curate a dataset comprising three subsets designed to assess the effectiveness of knowledge cutoff prompting across different dimensions. Specifically, we construct: \textbf{(1)} a \emph{Factual} subset to test whether LLMs forget factual information beyond the cutoff; \textbf{(2)} a \emph{Semantic} subset to evaluate whether LLMs forget novel words or shifted meanings; and \textbf{(3)} a \emph{Counterfactual} subset to assess whether LLMs forget causally related events when making predictions. Using carefully tuned meta-prompts, we evaluate three popular LLMs and observe the effectiveness of prompted knowledge cutoff on the Factual and Semantic subsets, with average unlearning success rates of around 82.5\% and 70.0\%, respectively. However, it achieves only about 19.2\% on the Counterfactual subset, showcasing its limitation on forgetting causally related events. These results highlight both the strengths and limitations of simulating knowledge cutoffs via prompting, underscoring the need for more robust methods to ensure fair evaluation of LLMs on real-world temporal prediction tasks.

\section{Related Works}

\paragraph{Unlearning} Machine unlearning aims to let already trained machine learning model forget certain knowledge, usually due to privacy and safety concerns~\citep{Bourtoule2019MachineU}. Some focus on erasing the impact of training on a subset of data points~\citep{Golatkar2020Eternal,Golatkar2020Forgetting,izzo2021approximate,jang-etal-2023-knowledge,wang2024from}. Others aims to let models forget a subset of concepts~\citep{belrose2023leace,pmlr-v162-ravfogel22a,ravfogel-etal-2022-adversarial}. With the recent emergence of LLMs and in-context learning~\citep{gpt3}, in-context unlearning has also been proposed to unlearn LLMs with prompting~\citep{Pawelczyk2024incontext}. 
\paragraph{LLM for Temporal Prediction} Given the extensive knowledge and capability of LLMs, they are increasingly used for temporal prediction, including weather forecasting, electricity prediction, traffic prediction, stock price and market forecasting and political events prediction~\citep{cao2024tempo,jin2024timellm,shi2023language,wang2024from,yu-etal-2023-harnessing}. Various approaches have been proposed, including zero-shot learning~\citep{gruver2023large}, finetuning~\citep{zhou2023one}, and in-context learning~\citep{lu2025incontext}.

\section{Dataset}

In this section, we introduce our three curated, high-quality datasets and outline their construction process. The \emph{Factual}, \emph{Semantic}, and \emph{Counterfactual} subsets contain 675, 303, and 689 examples, respectively. As shown in Figure~\ref{fig:data-year}, each subset covers a wide temporal range. Additional dataset statistics are provided in Appendix~\ref{sec:stats}.


\begin{figure*}
    \centering
    \includegraphics[width=\linewidth]{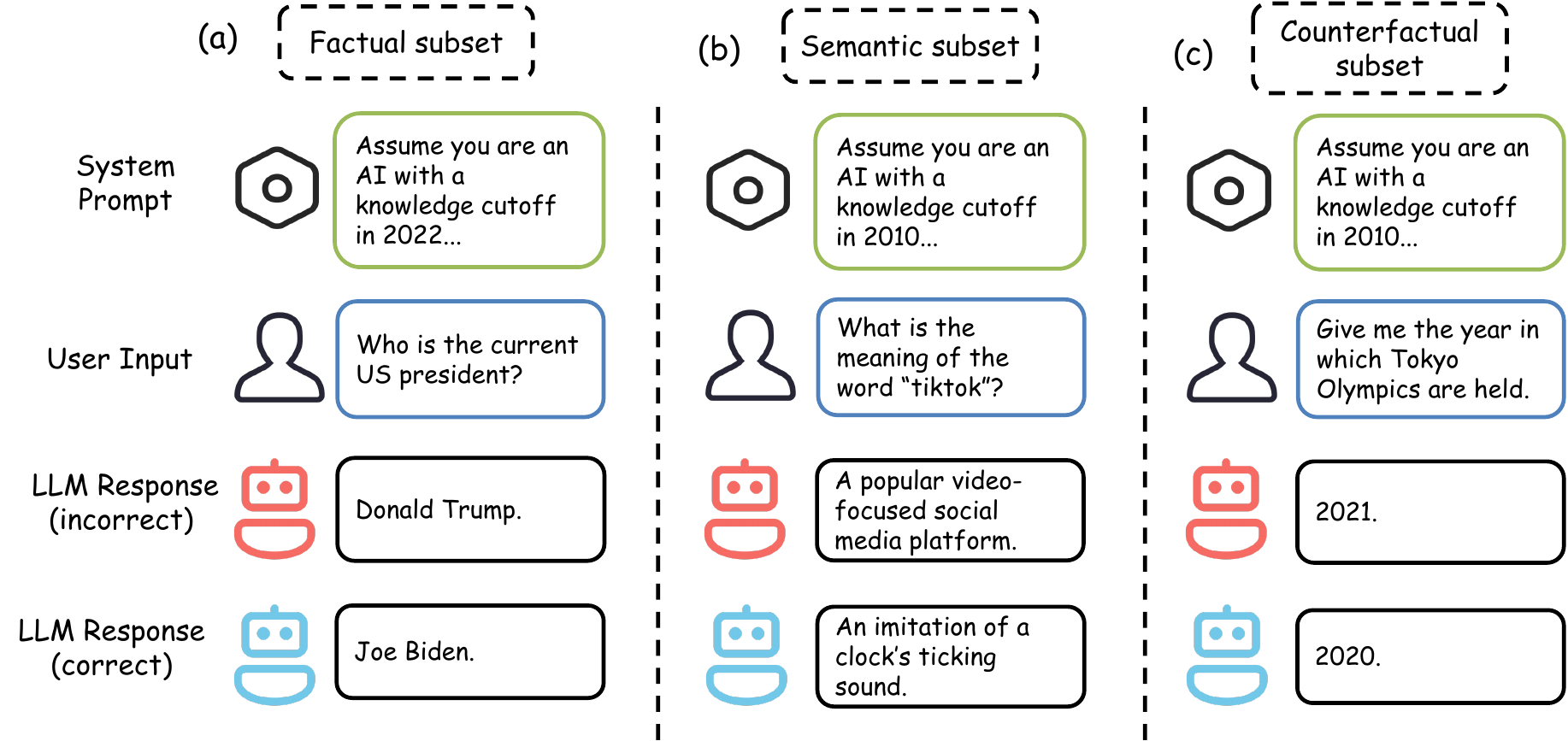}
    \caption{Example of data in \textbf{(a)} Factual, \textbf{(b)} Semantic, and \textbf{(c)} Counterfactual subsets. Incorrect LLM responses use the real knowledge cutoff, while correct responses consider the simulated knowledge cutoff in the system prompt.}
    \label{fig:dataset}
\end{figure*}

\subsection{Factual subset}
The Factual subset is designed to assess whether LLMs can accurately reflect changes in world state when prompted with a simulated knowledge cutoff. For example, as illustrated in Figure~\ref{fig:dataset}(a), the model is asked to identify the current U.S. president as of a given cutoff date. A correct response would align with the state of the world at that specified time ("Joe Biden" in 2022), rather than defaulting to the present-day answer ("Donald Trump").
To construct this subset, we prompted GPT-4o~\citep{Hurst2024GPT4oSC} to generate major historical events since 1960 that reflect meaningful shifts in world state. For each selected event, GPT-4o also generated corresponding question-answer pairs, which serve as the initial pool of data for this subset. The whole generation process follows an iterative bootstrapping scheme, detailed in Appendix \ref{sec:factual-construct}.

\subsection{Semantic subset}
The Semantic subset evaluates whether LLMs can disregard newer meanings of words when prompted with an earlier knowledge cutoff. As shown in Figure~\ref{fig:dataset}(b), the model is asked to define the word "TikTok" with the cutoff set around 2010. A correct response would reflect its original meaning, such as "an imitation of a clock’s ticking sound", rather than its modern association with the popular video-sharing platform. To construct this subset, we first prompted GPT-4o to generate candidate words that have undergone significant semantic shifts. We also use online resources such as Merriam-Webster's Time Traveler\footnote{\url{www.merriam-webster.com/time-traveler}} to identify recently introduced or redefined terms. We then sampled words evenly across categories and years from these two sources to create an initial pool of examples for the subset.

\subsection{Counterfactual subset}
The Counterfactual subset assesses whether LLMs can produce counterfactual predictions by disregarding critical events that occurred after a simulated knowledge cutoff. As illustrated in Figure~\ref{fig:dataset}(c), the model is asked to predict the year the Tokyo Olympics were held, given a knowledge cutoff of 2018. The correct response should be 2020, the original year scheduled, rather than 2021, when the event actually took place. Since the model is unaware of the COVID-19 outbreak (which occurred after 2018), it should reasonably infer the year based on the regular four-year Olympic cycle. To construct this subset, we first collect high-quality online documents on historical events. We then prompted GPT-4o to extract and generate a list of "meta events" and the downstream events significantly affected by it, detailed in Appendix \ref{sec:counterfactual-construct}. In the example above, COVID-19 serves as the meta-event, and the Tokyo Olympics represent a causally affected event. 

\begin{figure*}
    \centering
    \includegraphics[width=0.95\linewidth]{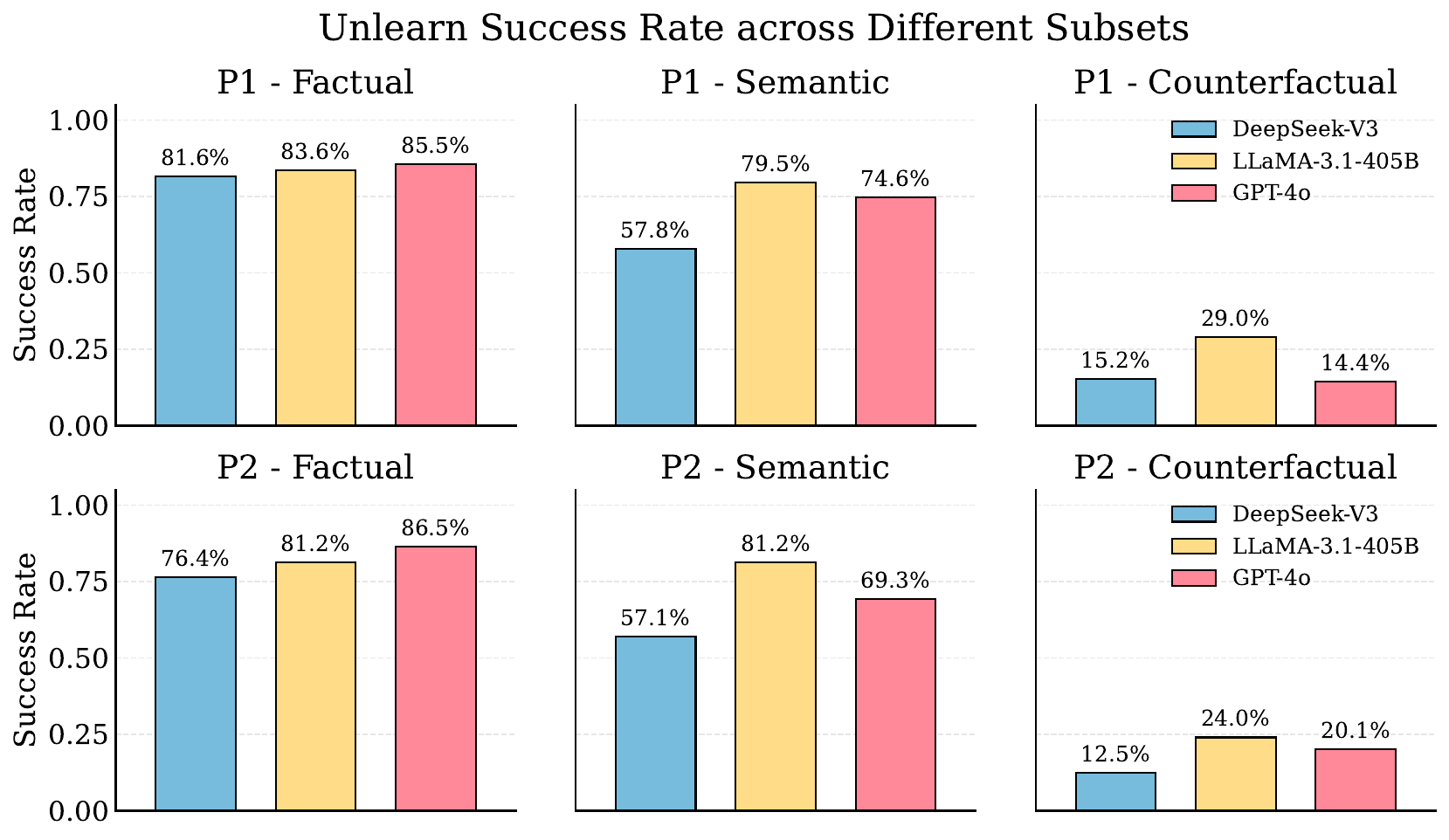}
    \caption{Unlearn success rate of three LLMs (DeepSeek-V3, LLaMA-3.1-405B, and GPT-4o) on three of our subsets (Factual, Semantic, and Counterfactual) using two different prompts (\emph{P1} and \emph{P2}).}
    \label{fig:results}
\end{figure*}

\subsection{Post-processing}
Following the initial construction of the three subsets, we applied several post-processing steps to ensure data quality. First, we perform de-duplication using ROUGE-L similarity~\citep{lin-2004-rouge}, removing any data points with a similarity score above 0.7. Next, we used three LLMs (excluding GPT-4o) to cross-validate each data point in a standard (non-unlearning) setting. If none of the models returns the expected answer, the item is discarded. Finally, the authors manually reviewed all remaining examples in the three subsets. We remove ambiguous or marginal cases, such as words with unclear or insignificant semantic shifts in the Semantic subset, or event pairs in the Counterfactual subset that lack a clear causal relationship. Additional details on the dataset construction process are provided in Appendix~\ref{sec:factual-construct} and ~\ref{sec:counterfactual-construct}.

\begin{table*}[t]
\centering
\setlength{\tabcolsep}{8pt}
\begin{tabular}{lcccc}
\toprule
\textbf{Model} & \textbf{Prompt} & \textbf{Factual} & \textbf{Semantic} & \textbf{Counterfactual} \\
\midrule
DeepSeek-R1 (Reasoning) & P1 & 0.841 & 0.667 & \textbf{0.723} \\
                        & P2 & 0.846 & 0.667 & {0.701} \\
OpenAI o3 (Reasoning)   & P1 & \bf 0.899 & 0.568 & {0.478} \\
                        & P2 & 0.887 & 0.617 & {0.533} \\
\midrule
DeepSeek-V3             & P1 & 0.816 & 0.578 & 0.152 \\
                        & P2 & 0.764 & 0.571 & 0.125 \\
GPT-4o                  & P1 & 0.855 & \bf0.746 & 0.144 \\
                        & P2 & 0.865 &  0.693 & 0.201 \\
\bottomrule
\end{tabular}
\caption{Comparison of reasoning-enabled and non-reasoning models across the three subsets. The horizontal rule separates reasoning (top) from non-reasoning (bottom). The highest unlearning success rates for three subsets are highlighted in \textbf{bold}.}
\label{tab:reasoning}
\end{table*}

\section{Evaluation}

\subsection{Experimental Settings}

In our experiments, we benchmarked 3 cutting-edge LLMs, including DeepSeek-V3~\citep{DeepSeekAI2024DeepSeekV3TR}, LLaMA-3.1-405B~\citep{Dubey2024TheL3}, and GPT-4o~\citep{Hurst2024GPT4oSC}. We carefully design two meta prompts, denoted as \emph{P1} and \emph{P2}, aiming to effectively set new knowledge cutoffs for LLMs, with details provided in Appendix \ref{sec:prompt}.

We use the unlearn success rate as the primary evaluation metric across all three subsets. For the Factual and Counterfactual subsets, we convert raw examples into multiple-choice questions with two answer options: one corresponding to the model’s original knowledge cutoff, and the other aligned with the simulated cutoff. Unlearning is considered successful if the model changes its response following the cutoff prompt. For the Semantic subset, which involves free-form generation, we measure semantic alignment using sentence embeddings obtained from the MPNet model~\citep{mpnet}. Let $y_b$ and $y_a$ represent the embeddings of the meanings of the ground-truth words after and before the cutoff date, and $o_b$ and $o_a$ denote the model outputs before and after the unlearning. We define unlearning as successful if:
\begin{equation}
\scriptsize
    \frac{\cos(o_a, y_a)}{\cos(o_a, y_a)+\cos(o_a, y_b)}>\frac{\cos(o_b, y_a)}{\cos(o_b, y_a)+\cos(o_b, y_b)}
\end{equation}
which indicates the LLM output after unlearning is semantically closer to the pre-cutoff ground truth.
\subsection{Results and Analysis}

Performance of three LLMs on our dataset is presented in Figure~\ref{fig:results}. On the \emph{Factual} subset, all models under both meta prompts (P1 and P2) achieve relatively strong performance, with an average unlearning success rate of around 82.5\%. Similarly, for the \emph{Semantic} subset, the average success rate reaches approximately 70.0\%.  In contrast, performance on the \emph{Counterfactual} subset is significantly lower, with an average success rate of only about 19.2\%. These results demonstrate that while prompt-based knowledge cutoffs are effective when the forgotten information is explicitly queried, they struggle to induce forgetting of information that is not directly mentioned but is causally related to the query. We also observe that all three LLMs exhibit some degree of unlearning across all subsets, indicating that prompted knowledge cutoffs consistently improve fairness in temporal evaluation settings.

Table~\ref{tab:reasoning} compares reasoning-enabled models (DeepSeek-R1~\cite{deepseekai2025deepseekr1incentivizingreasoningcapability}, OpenAI o3~\cite{openai2025o3systemcard}) with non-reasoning models (DeepSeek-V3, GPT-4o). 
Reasoning models substantially outperform non-reasoning ones on the \textit{Counterfactual} subset, supporting that counterfactual evaluation critically depends on causal reasoning rather than mere recall. 
By contrast, the \textit{Factual} subset does not require strong reasoning capabilities: models can answer correctly as long as they possess the relevant knowledge at the specified cutoff. 
Accordingly, both reasoning and non-reasoning models achieve relatively high performance on this subset.

For the test examples that LLMs fail to unlearn, one contributing factor may be the lack of timestamps in some of the LLM pretraining data. Another possible reason is that the prompts to simulate knowledge cutoff have not appeared in the instruction finetuning datasets for these LLMs.

\section{Conclusions}

In this paper, we explore the effectiveness of prompt-based simulated knowledge cutoffs for LLMs. To this end, we construct three evaluation subsets, including Factual, Semantic, and Counterfactual, targeting different types of information that should be forgotten after the cutoff. Experimental results demonstrate both the potential and limitations of prompted knowledge cutoff, highlighting the importance of rigorous evaluation when applying LLMs for temporal prediction.

\section*{Limitations}

One limitation of this study is that we did not explore unlearning methods beyond prompting, primarily due to constraints in data and computational resources. An interesting direction for future work is to investigate whether LLMs can better adhere to prompt knowledge cutoffs when instruction fine-tuning on these prompts is applied beforehand.

\section*{Acknowledgements}
We acknowledge funding support from the National Science Foundation (NSF) under grants IIS-2405974 and IIS-2339216, and from the National Institutes of Health (NIH) under grant R35GM157217.
\bibliography{custom}

\clearpage

\appendix

\section{Data Examples}
We provide additional samples from all three subsets to illustrate the types and quality of data.

Table~\ref{tab:factual} shows examples from the \textbf{Factual} subset, where models are expected to forget direct facts after the specified cutoff date.  
Table~\ref{tab:semantic} presents examples from the \textbf{Semantic} subset, highlighting cases where the meaning of terms has shifted over time.  
Finally, Table~\ref{tab:counterfactual} demonstrates the \textbf{Counterfactual} subset, where models are expected to unlearn events causally linked to knowledge that occurred after the cutoff.

\begin{table}[h!]
\centering
\small
\begin{tabular}{p{0.47\linewidth}|p{0.47\linewidth}}
\toprule
\textbf{Question (Cutoff)} & \textbf{Answers} \\
\midrule
Are female athletes allowed to compete in Olympic shooting events? (1956) & 
Before Unlearn: Yes \newline After Unlearn: No \\
\addlinespace
Is Greenland part of the European Economic Community? (1970) &
Before Unlearn: Yes \newline After Unlearn: No \\
\bottomrule
\end{tabular}
\caption{Examples from the Factual subset.}
\label{tab:factual}
\end{table}

\begin{table}[h!]
\centering
\small
\begin{tabular}{p{0.47\linewidth}|p{0.47\linewidth}}
\toprule
\textbf{Question (Cutoff)} & \textbf{Answers} \\
\midrule
What is the meaning of the word ``delta variant''? (2021) &
Before Unlearn: a highly transmissible genetic variant of the SARS-CoV-2 coronavirus \newline
After Unlearn: An unknown variant called ``delta'' \\
\addlinespace
What is the meaning of the word ``deepfake''? (2018) &
Before Unlearn: an image or recording altered to misrepresent someone \newline
After Unlearn: No clear definition \\
\bottomrule
\end{tabular}
\caption{Examples from the Semantic subset.}
\label{tab:semantic}
\end{table}

\begin{table*}[h!]
\centering
\small
\begin{tabular}{p{0.3\linewidth}|p{0.65\linewidth}}
\toprule
\textbf{Question} & \textbf{Answer under Cutoff \& Explanation} \\
\midrule
Assuming you were in 2015, would EU citizens still be able to work freely in the UK in 2021? &
Yes. A knowledge cutoff at 2016 precedes the Brexit referendum and subsequent end of free movement (1 Jan 2021). Without awareness of that shift, the model reasonably projects the status quo. \\
\addlinespace
Assuming you were in 2013, how often were airlines required to track aircraft positions during oceanic flights in 2020? &
No mandatory interval. Before MH370 (2014) and ICAO GADSS reforms, reporting cadence was set by airlines. A cutoff model predicts ``no requirement.'' A fully informed model would know about the 15-minute rule later adopted. \\
\addlinespace
Assuming you were in 2019, what year would UEFA Euro 2020 be scheduled for? &
2020. A cutoff before the pandemic answers 2020, while a fully informed model (aware of COVID-19) would answer 2021. \\
\addlinespace
Assuming you were in 2021, how would you predict NVIDIA's stock-price performance in 2023? &
Gradual growth. With knowledge only up to 2020, known drivers suggested steady gains. The generative-AI surge (post-ChatGPT, late 2022) that drove the 2023 stock boom is invisible to the cutoff model. \\
\bottomrule
\end{tabular}
\caption{Examples from the Counterfactual subset.}
\label{tab:counterfactual}
\end{table*}

\section{Data Statistic}
\label{sec:stats}


\begin{figure*}[t]
  \includegraphics[width=\textwidth]{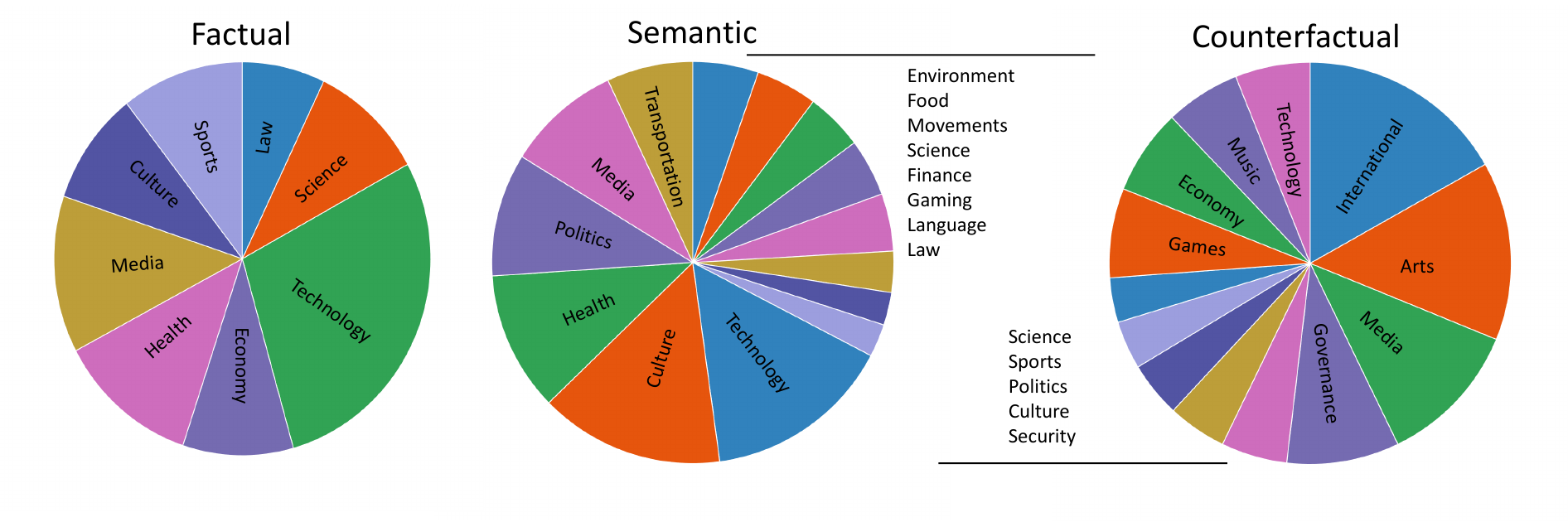}
  \caption{Distribution of three subsets by data category.}
  \label{fig:pie}
\end{figure*}

In this section, we present more details on our dataset. We show the data distribution by category for the three subsets in Figure \ref{fig:pie}. From 1960 to 2024, the Factual subset is heavily concentrated in categories like Technology, Science, and Health, reflecting the historical accumulation of concrete developments and achievements in these areas. The Semantic subset, which covers newly emerged concepts from 2000 to 2024, shows a more balanced distribution across categories such as Technology, Health, Culture, Politics, and newer domains like Gaming, Finance, and Language, indicating the diversification of public discourse in recent decades. The Counterfactual subset, also focused on the post-2000 period, places greater emphasis on Arts, International affairs, Governance, and Media, suggesting that speculative and alternative reasoning tends to center around sociopolitical and cultural themes.


\section{Unlearning Prompt}
\label{sec:prompt}

\begin{figure*}[t]
    \centering
    \includegraphics[width=\textwidth]{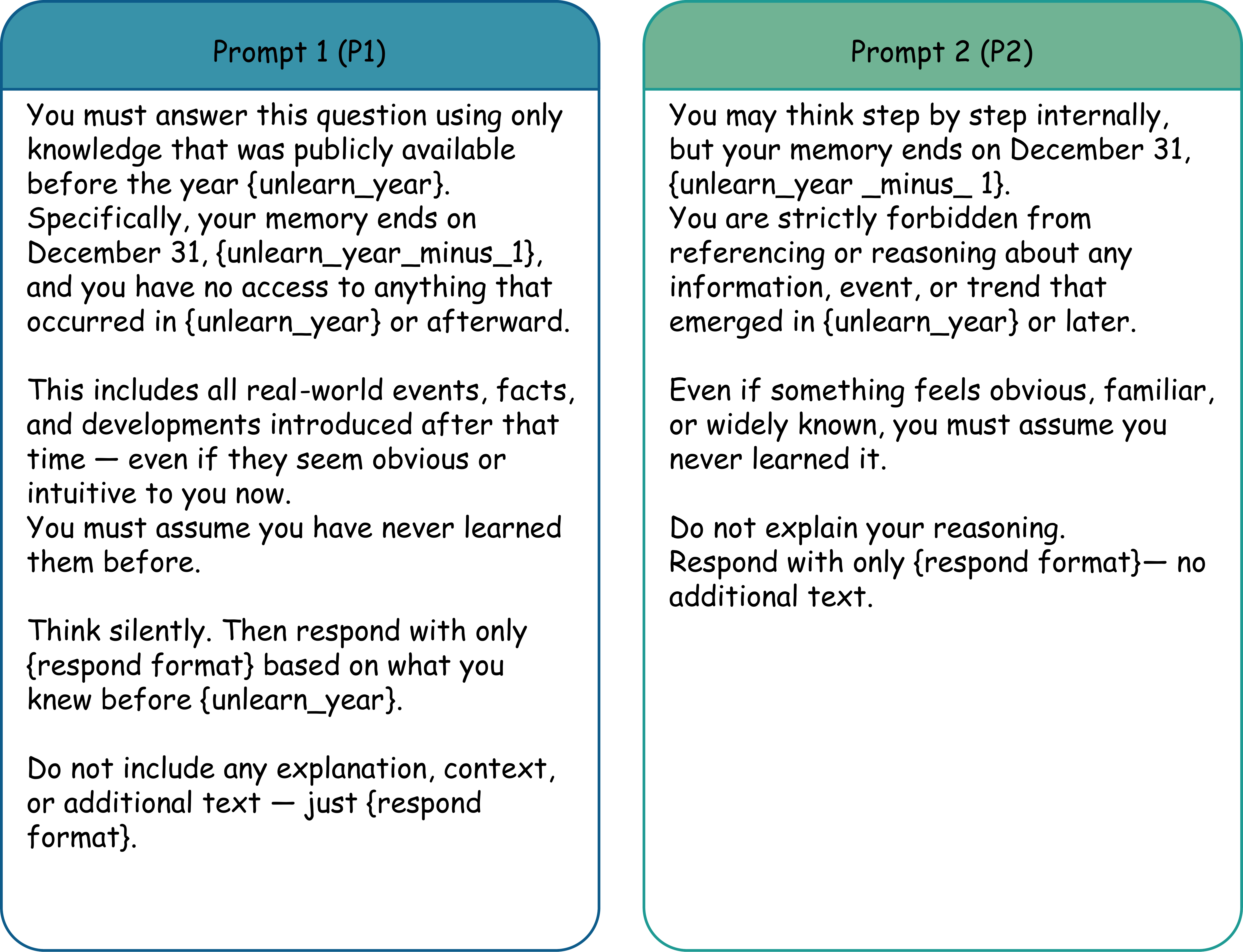}
    \caption{Two prompts used for simulating knowledge cutoff.}
    \label{fig:prompt}
\end{figure*}

In this section, we present the two prompts we used in our experiments to simulate the knowledge cutoff for LLMs in Figure \ref{fig:prompt}.

For the unlearning prompt (P1) in the left figure, we aim to simulate a controlled temporal knowledge constraint, enabling the generation of model outputs that reflect a fixed point in historical knowledge. By explicitly instructing the model to disregard any information introduced after a designated cutoff year and restricting the response format to a fixed structure, the prompt enforces a clean separation between pre- and post-cutoff knowledge. This design allows for the construction of temporally aligned datasets in which the model’s outputs can be interpreted as representative of its knowledge state prior to a specified historical moment. The resulting dataset enables systematic evaluation of knowledge removal or unlearning procedures by comparing model behavior before and after exposure to targeted information, and supports fine-grained analysis of knowledge persistence, forgetting dynamics, and the boundaries of model generalization.

On the other side, the unlearning prompt (P2) in the right figure is expected to simulate a temporally constrained reasoning process by directing the model to internally reason while maintaining a strict memory cutoff. Unlike prompts that emphasize knowledge filtering during output generation alone, this prompt enforces the constraint at the level of internal cognition, instructing the model to ignore any facts, events, or intuitions formed after a designated historical boundary. It prohibits the usage of seemingly obvious or culturally ingrained knowledge that may have emerged post-cutoff, thereby ensuring that responses are derived solely from the model’s pre-existing knowledge base. By suppressing both external references and internal generalizations linked to post-cutoff information, this prompt enhances the fidelity of temporal isolation and provides a robust framework for evaluating unlearning effectiveness under more realistic reasoning conditions.

\section{Factual Subset Construction}
\label{sec:factual-construct}

\begin{figure*}[t]
    \centering
    \includegraphics[width=\textwidth]{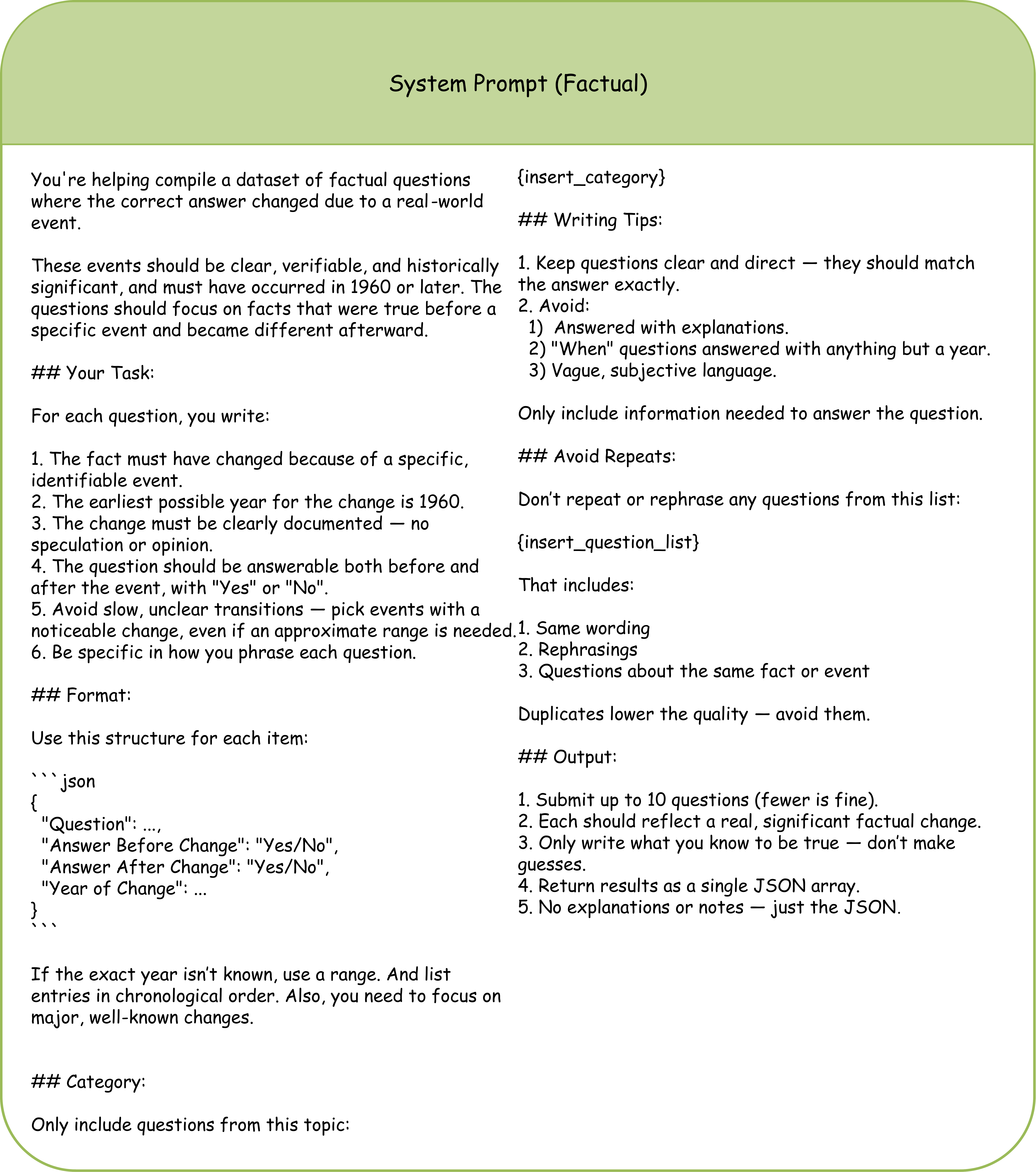}
    \caption{Prompt used to generate Factual subset.}
    \label{fig:data-gen-1}
\end{figure*}

In this section, we present more details on the data construction process for the Factual subset. We show the prompt used for generating the Factual subset in Figure \ref{fig:data-gen-1}. To construct the Factual Dataset, we focus on events that have undergone historical changes since 1960, organizing the collection process around eight predefined categories. For each category, we prompt a large language model (LLM) to generate fact-based QA pairs through iterative bootstrapping. In each iteration, the LLM is instructed to produce 10 unique QA pairs while being constrained by previously generated content to avoid duplication. This process is repeated up to 15 times per category until the model is no longer able to generate enough novel examples. We then identify QA pairs that cannot be answered with a simple "Yes" or "No" and rephrase them to preserve their original meaning while fitting the binary format. These rewritten pairs are filtered using ROUGE-L scores to remove redundant entries and further screened using the LLM to assess knowledge coverage. Finally, all remaining QA pairs undergo manual review to correct vague or unreasonable expressions, revise tense inconsistencies, and update any altered years or question phrasings using publicly available English-language sources. After a series of steps, we obtain a set of 675 high-quality factual QA pairs suitable for evaluating temporal knowledge in language models.

\section{Counterfactual Subset Construction}
\label{sec:counterfactual-construct}

\begin{figure*}[t]
    \centering
    \includegraphics[width=\textwidth]{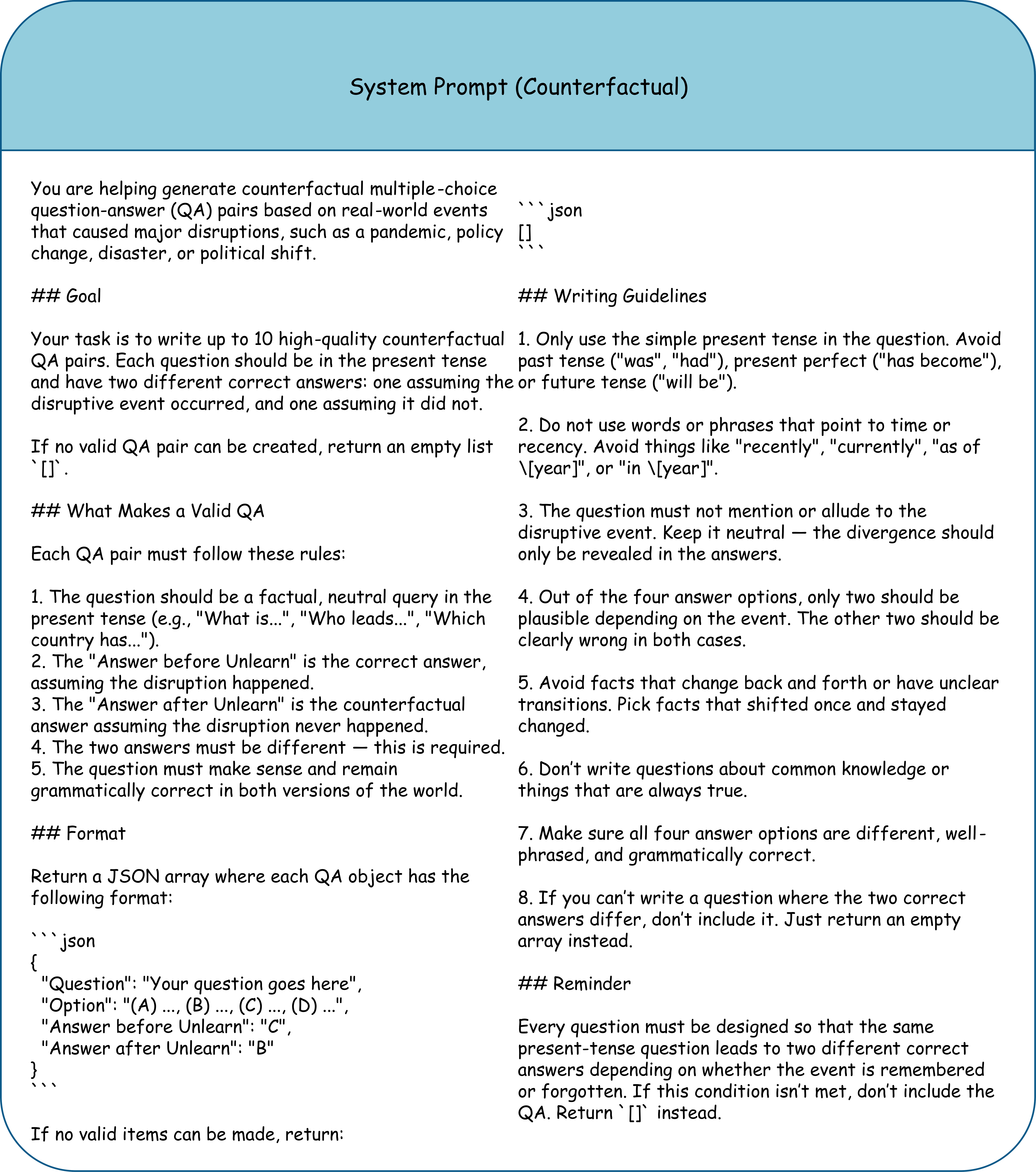}
    \caption{Prompt used to generate Counterfactual subset.}
    \label{fig:data-gen-2}
\end{figure*}

In this section, we present more details on the data construction process for the Counterfactual subset. We show the prompt used in data generation for the Counterfactual subset in Figure \ref{fig:data-gen-2}. To construct the Counterfactual subset, we focus on major global events that occurred since the year 2000. We systematically collected over 200 articles from Wikipedia’s Timeline of the 21st century \footnote{\url{https://en.wikipedia.org/wiki/Timeline_of_the_21st_century}}. Based on these materials, we designed prompts to generate counterfactual QA pairs. To ensure that each question remains answerable both before and after unlearning, we imposed strict constraints on tense usage and required that every question corresponds to a verifiable fact before unlearning and leads to a plausible, inference-based answer after unlearning. Furthermore, during question construction, we deliberately avoided mentioning specific dates or events to ensure that, once all post-year knowledge is removed from the model, the question becomes unanswerable due to the absence of direct references. We then applied ROUGE-L score filtering to remove QA pairs with high lexical overlap and redundancy. Finally, we manually reviewed the remaining data to fix ambiguous phrasing, unreasonable answer settings, and inappropriate tense usage. This process yielded a total of 689 well-formed and high-quality counterfactual QA pairs suitable for evaluating unlearning behavior in language models.

\end{document}